\documentclass[a4paper,conference]{IEEEtran}
\usepackage{graphicx}
\usepackage{multirow}
\usepackage{array}
\usepackage{float}
\usepackage{amsmath,bm}
\usepackage{algorithm}
\usepackage{algpseudocode}
\usepackage{pgfplots}
\usepackage{pgfplotstable}
\usepackage{slashbox}
\usepackage{caption}
\usepackage{amssymb}
\usepackage{arydshln}
\usepackage{subcaption}
\usepackage{xcolor}
\usepackage{url}
\usepackage{stackengine}

\usepackage[T1]{fontenc}
\usepackage[utf8]{inputenc}

\newcommand\IncG[2][]{\addstackgap{%
  \raisebox{-.5\height}{\includegraphics[#1]{#2}}}}

\newcommand{\ResidualLayerNumber}{\beta}

\begin{document}

%
\title{Kunster - AR Art Video Maker - Real time video neural style transfer on mobile devices}

\author{
\IEEEauthorblockN{Wojciech Dudzik, Damian Kosowski}
\IEEEauthorblockA{Netguru S.A., \\ul. Małe Garbary 9
 61-756 Poznań, Poland \\
Email: wojciech.dudzik@netguru.com, damian.kosowski@netguru.com}
 
}

\maketitle

\begin{abstract}
Neural style transfer is a well-known branch of deep learning research, with many interesting works and two major drawbacks. Most of the works in the field are hard to use by non-expert users and substantial hardware resources are required. In this work, we present a solution to both of these problems. We have applied neural style transfer to real-time video (over 25 frames per second), which is capable of running on mobile devices. We also investigate the works on achieving temporal coherence and present the idea of fine-tuning, already trained models, to achieve stable video. What is more, we also analyze the impact of the common deep neural network  architecture on the performance of mobile devices with regard to number of layers and filters present. In the experiment section we present the results of our work with respect to the iOS devices and discuss the problems present in current Android devices as well as future possibilities. At the end we present the qualitative results of stylization and quantitative results of performance tested on the iPhone 11 Pro and iPhone 6s. The presented work is incorporated in Kunster - AR Art Video Maker application available in the Apple's App Store.
\end{abstract}


\section{Introduction}

Painting as a form of art has accompanied us through the history, presenting all sorts of things, from the mighty kings portraits through, historic battles to ordinary daily activities. It all changed with the invention of photography, and later,  digital photography. Nowadays, most of us carry a smart phone equipped with an HD camera. In the past, re-drawing an image in a particular style required a well-trained artist and lots of time (and money). 

This problem has been studied by both artists and computer science researchers for over two decades within the field of non-photorealistic rendering (NPR). However, most of these NPR stylization algorithms were designed for particular artistic styles and could not be easily extended to other styles. A common limitation of these methods is that they only use low-level image features and often fail to capture image structures effectively. The first to use convolution neural networks (CNN) for that task was Gatys et al. \cite{Gatys2016ImageST,GatysEB15archivix}. They proposed a neural algorithm for automatic image style transfer, which refines a random noise to a stylized image iteratively constrained by a content loss and a style loss. This approach resulted in multiple works that attempted to improve the original and addressed its major drawbacks, such as: the long time needed to obtain stylization or applying this method to videos. In our work, we studied the possibility of delivering these neural style transfer (NST) methods for videos on mobile phones, especially the iPhone. In order to do this, we investigate the problem of temporal coherence among existing methods and propose another approach as we found problems with applying them to mobile devices. We also refined the neural network architecture with regard to its size; therefore, our main contributions are as follow:

\begin{itemize}
\item Real-time application of neural style transfer to videos on mobile devices (iOS).
\item Investigation into achieving temporal coherence with existing methods.
\item Analyzing the size of the models present in literature, and proposing a new smaller architecture for the task.
\end{itemize}


First we shall review the current status of the NST field related to image and video style transfer (see \ref{related-work}). Further in (section \ref{sec:methods}) we describe the training regime and proposed neural network architecture.  And finally, achieving temporal coherence will be presented in Section \ref{sec:temporal_coherence}. In Section \ref{sec:experiments}, we will discuss the results obtained during our experiment and show performance on mobile devices.


\section{Related work}
\label{related-work}

In this section, we will briefly review the selected methods for NST related to our work, while for a more comprehensive review, we recommend \cite{NeuralStyleTransferReview}.  The first method for NST was proposed by Gatys et al. \cite{Gatys2016ImageST}. He demonstrated exciting results that caught eyes in both academia and industry. That method opened many new possibilities and attracted the attention of other researchers, i.e.: \cite{JohnsonAL16,DumoulinSK16,Champandard16,LuanPSB17,GatysEBHS16}, whose work is based on Gatys original idea. One of the best successory projects was proposed by Johnson et al. \cite{JohnsonAL16} with the feed-forward perceptual losses model. In his work, Johnson used a pre-trained VGG \cite{Simonyan15} to compute the content and style loss. This allowed real-time inference speed while maintaining good style quality.  A natural way to extend this image processing technique to videos is to perform a certain image transformation frame by frame. However, this scheme inevitably brings temporal inconsistencies and thus causes severe flicker artifacts for the methods that consider single image transformation. One of the methods that solved this issue was proposed by Ruder et al. \cite{RuderDB16}, which was specifically designed for video. Despite its usability for video, it requires a time-consuming computations (dense optical flow calculation), and may take several minutes to process a single frame. Due to this fact, it makes it not applicable for real-time usage. To obtain a consistent and fast video style transfer method, some real-time or near real-time models have recently been developed.

Using a feed-forward network design, Huang et al. \cite{Huang_2017_CVPR} proposed a model similar to the Johnson's  \cite{JohnsonAL16} with an additional temporal loss. This model provides faster inference times since it neither estimates optical flows nor uses information about the previous frame at the inference stage. Another, more recent, development published by Gao et al. \cite{Reconet} describes a model that does not estimate optical flows but involves ground-truth optical flows only in loss calculation in the training stage. The use of the ground-truth optical flow allows us to obtain an occlusion mask. Masks are further used to mark pixels that are untraceable between frames and should not be included in temporal loss calculations. Additionally, temporal loss is considered not only on the output image but also at the feature level of DNN. Gao's lightweight and feed-forward network is considered to be one of the fastest approach for video NST.

Still, applying methods mentioned above may be troublesome due to limited capabilities of mobile devices. Even though modern smartphones are able to run many machine learning models, achieving real-time performance introduces more strict requirements about the model design. There was several noticeable reports dealing with this issue, i.e. \cite{TinyTransformer,iosApp,Pictory}. In these papers, authors are focusing on running current methods such as Johnson \cite{JohnsonAL16}, on mobile platforms. In order to meet the desired performance authors are encompassing the use of specialized ARM instructions - NEON or GPU computation while all of them perform only image to image style transfer. Other implementations include \cite{Novecento} and very popular Prisma application. Both of them relay on processing images on server side (although Prisma added later option for on device processing). As a consequence, both of them heavily depend on the internet connection and enable processing of single images at the same time.

Since the pace of progression in the hardware capabilities (CPU, GPU) of mobile devices is very fast, the computation power grows each year. This trend was shown clearly in Ignatov et al\cite{aiMobileBenchmark}, where the authors present a comprehensive review of smartphones performance against popular machine learning and deep learning technologies used nowadays. In particular, they analyzed devices with Android operating systems. The introduction of fourth generation (according to Ignatov's designations) neural processing units (NPU) for Android phones makes it easier for any deep neural network application as many limitations are removed by the newer hardware. Moreover, unifying programming interface with frameworks like Tensorflow Lite, Android NNAPI, or CoreML (for iPhones) ease the development process for any DNN application. Before that, deploying a single model to multiple devices was difficult since one would need to integrated it with multiple different SDKs prepared by each of chip manufacturer, e.g., MNN, Mace. Each of them also provided also certain limitations in term of supported operations, model quantization options, and hardware usage.

This motivated us to pursue possibilities for work involved in both areas of interest: NST and mobile applications of DNN. As a result, we propose improvements by combining the methods from those fields and propose a real-time video neural style transfer on mobile devices.

\section{Proposed Method}
\label{sec:methods}

We propose a reliable method for achieving real-time neural style transfer on mobile devices. In our approach, we are primarily focused on iPhones. There are still some differences between them and Android-based phones, which we will address in Section \ref{Andorid_and_ios}. In this section, we will present our network architecture and training procedure. 

\subsection{Network architecture}

\begin{figure*}[ht]
\caption{ReCoNet architecture. Symbols are as follow: $I_t$, $F_t$, $O_t$ are input image, encoded feature map and output image respectively at time point $t$. The $I_{t-1}$, $F_{t-1}$, $O_{t-1}$ are the results obtained in previous frame $t-1$. The $Style$ represents the artistic image used for stylization while $M_t$ and $W_t$ denote the occlusion mask and optical flow between frames $t-1$ and $t1$.  The loss function are denoted with red text \cite{Reconet}.}
\label{networkArchitecture}
\centering
\includegraphics[width=0.95\textwidth]{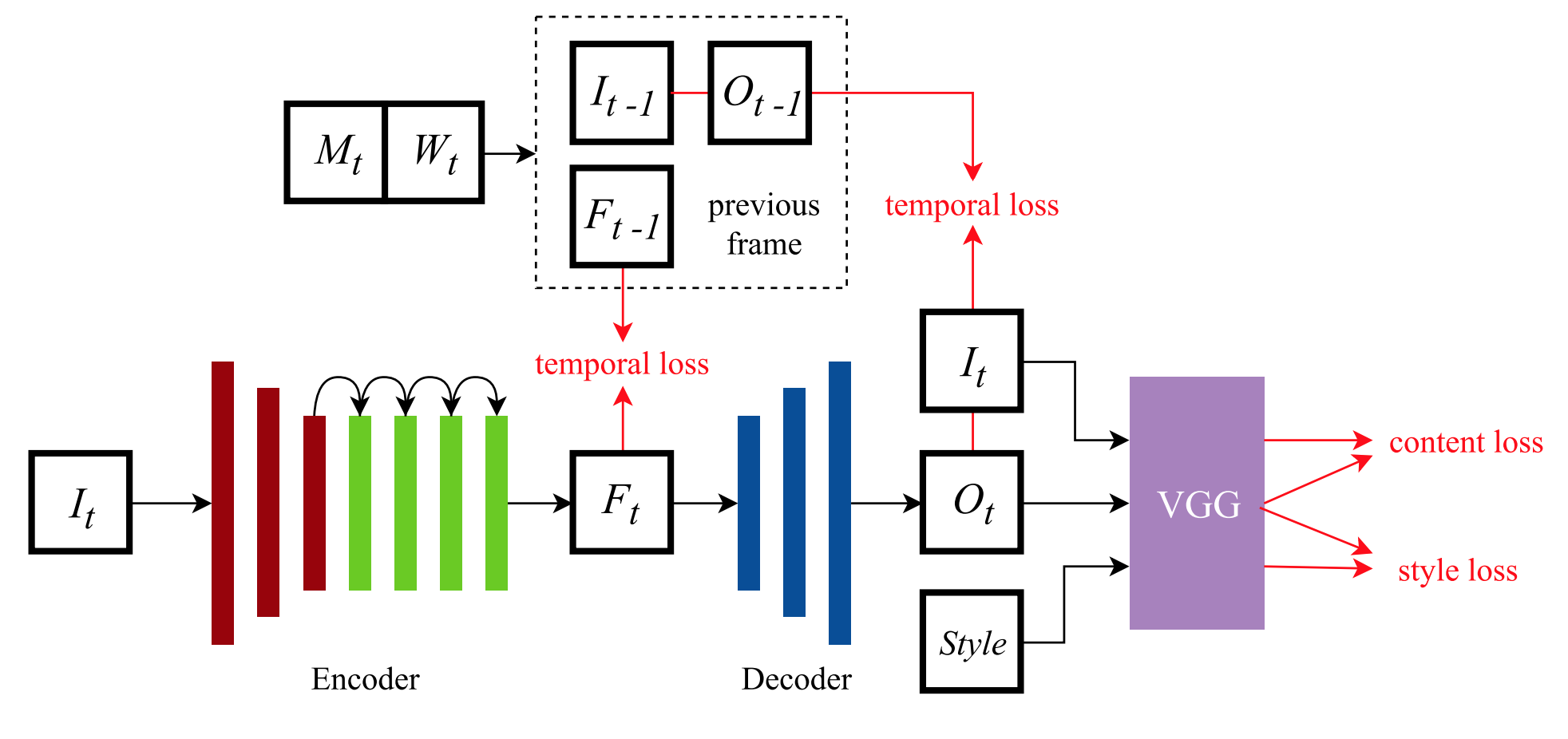}
\end{figure*}

In Fig. \ref{networkArchitecture}, we present the architecture of our network, which is the architecture present in \cite{Reconet}. In this paper, the architecture is composed of three main components: 1) the encoder, 2) the decoder, 3) the VGG-16 network. The encoder part is responsible for obtaining feature maps while the decoder generates stylized images from these feature maps. The last part of the VGG network is used for perceptual loss calculation. During the inference, only the encoder and decoder parts are used. A detailed description of each layer of the network was presented in Tab.  \ref{tableWithArchitecture}, with a focus on the number of filters.

\begin{figure}[ht]
\caption{Architecture of residual block}
\label{residualBlockArchitecture}
\centering
\includegraphics[scale=0.5]{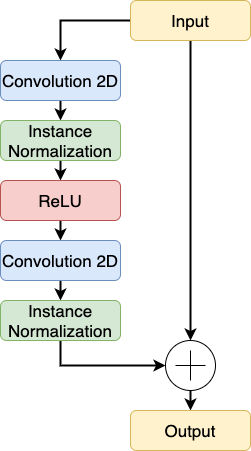}
\end{figure}

We provide  modifications to the original architecture, including changes in the number of filters at each layer (showed in last column of Tab. \ref{tableWithArchitecture}), and removing the TanH opertation at the last layer. Moreover, all of the kernel sizes were equal to $3 \times 3$ as opposed to \cite{Reconet} where first and last layer have kernel size of $9 \times 9$. We used reflection padding for each of the convolutions. For the upsample layer, we used the nearest neighbors method. A visualization of the residual layer architecture is presented in Fig. \ref{residualBlockArchitecture}. The $\ResidualLayerNumber$ is introduced in order to investigate the influence of the number of residual layers on the final result of stylization.

\begin{table}[ht]
\scriptsize
\caption{Detailed layer by layer architecture of network used}
\label{tableWithArchitecture}
\centering
\begin{tabular}{cccc}
\hline 
Layer & Type & Filters in \cite{Reconet} & Our filters \\ 
\hline 
1  & Input &  &  \\ 
\hline 
2 & Conv + instnorm + Relu & $ 48$  & $\alpha \cdot 32$ \\ 
\hline 
3 & Conv + instnorm + Relu (stride 2) & $96$ & $\alpha \cdot 48$ \\ 
\hline 
4& Conv + instnorm + Relu (stride 2) & $192$ & $\alpha \cdot 64$ \\ 
\hline 
5-8 & Residual  $\times  4 \cdot \ResidualLayerNumber$   & $192$ &  $\alpha \cdot 64$ \\ 
\hline
9 & Upsample & &\\
\hline 
10 & Conv + instnorm + Relu & $96$ & $\alpha \cdot 48$ \\ 
\hline 
11 & Upsample & &\\
\hline 
12 & Conv + instnorm + Relu & $48$ & $\alpha \cdot 32$ \\ 
\hline 
13 & Conv  & $3$ & $3$ \\ 
\hline 
\end{tabular} 
\end{table}

\subsection{Training procedure}
\label{sec:temporal_coherence}

As there is no code published by the authors of \cite{Reconet}, we tracked an open source implementation on the GitHub webpage\footnote{\url{https://github.com/EmptySamurai/pytorch-reconet}}. As far as we know, this implementation closely resembles the work presented in the paper but shows additional artifacts (in the form of white bubbles) that were not mentioned in original work. These phenomena are  presented in Fig. \ref{fig:problems}c, where artifacts can be easily seen. An interesting and detailed explanation of this problem was presented in \cite{karras2019analyzing}. The authors discovered that artifacts appear because of instance normalization employed to increase training speed. It is known that the latter is considered a better alternative than batch normalization \cite{instanceNormalization} for style transfer tasks. The solution proposed by the author of this repository is to use filter response normalization \cite{FRNlayer}. All instance normalization layers are replaced by filter response normalization (FRN) and all ReLU operations are replaced by a thresholded linear unit (TLU). Despite the fact that this solution removes problematic artifacts, it still  cannot be deployed on mobile devices, because FRN is not yet supported by any major framework at the time of writing. Of course, it can be computed on the general CPU, but this results in a major performance drop. On the other hand, our implementation of ReCoNet, in particular cases may present faded colors, as seen in Fig. \ref{fig:problems}b, and the final effect may not be as good as expected (\ref{fig:problems}a). The result presented in Fig. \ref{fig:problems}a was achieved by training the same architecture only with content and style loss. This schema reassembles the work of \cite{JohnsonAL16}. In our opinion, there might be some implementation details in ReCoNet  that we are missing, which are crucial for reaching the desired result. While we were not able to reproduce ReCoNet work or hit the obstacles with deployment on mobile devices, we came up with the idea of two-stage training as follows:

\begin{figure*}[ht]
\begin{tabular}{ccc}
\IncG[width=.3\textwidth]{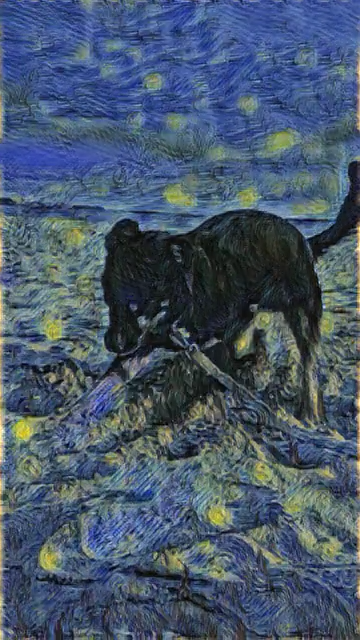}  &
\IncG[width=.3\textwidth]{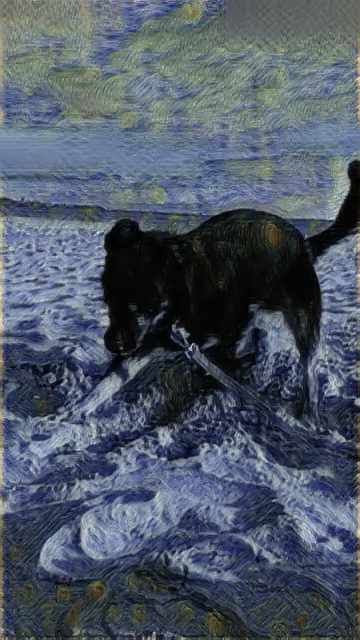} &
\IncG[width=.3\textwidth]{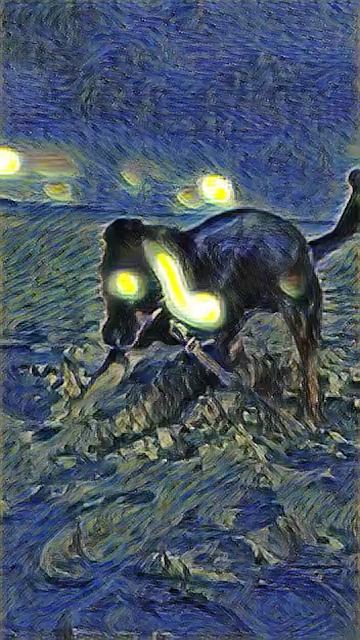}   
\\
a) Expected & b) Bad colors & c) Artifacts 
\end{tabular}
\caption{Problems that turned out on achieving temporal coherence}
\label{fig:problems}
\end{figure*}

\begin{enumerate}
\item Training only with content and style loss, as in \cite{JohnsonAL16}.
\item Fine-tuning the network with feature and output temporal losses added (losses came from  \cite{Reconet}).
\end{enumerate}

During the first stage of the training we adopt the content loss $\mathcal{L}_{content}$ and the style loss  $\mathcal{L}_{style}$ and the total variation regularizer  $\mathcal{L}_{tv}$ in the perceptual losses model based on \cite{JohnsonAL16}. The final form of the loss function for the first stage is:

\begin{equation}
 \mathcal{L} (t) =  \gamma \mathcal{L}_{content} + \rho \mathcal{L}_{style} + \tau \mathcal{L}_{tv}
\end{equation}

During the second stage of fine-tuning and achieving temporal consistency we adopt the solution presented in \cite{Reconet}, where  

\begin{equation}
\begin{aligned}
 \mathcal{L} (t-1,t) =  \sum_{i \in \{t-1, t\}} \left( \gamma \mathcal{L}_{content}(i) + \rho \mathcal{L}_{style}(i)  + \tau \mathcal{L}_{tv}(i) \right) \\ + \lambda_{f} \mathcal{L}_{temp,f}(t-1,t) + \lambda_{o} \mathcal{L}_{temp,o}(t-1,t)
 \end{aligned}
\end{equation}

where $\lambda_{f} \mathcal{L}_{temp,f}(t-1,t) + \lambda_{o} \mathcal{L}_{temp,o}(t-1,t)$ are feature temporal and output temporal loss components respectively, as presented in \cite{Reconet}. 

The key idea in achieving temporal consistency is the usage of optical flow information between the consecutive frames and occlusion masks for marking untraceable pixels, which should be excluded from the loss calculation. By doing this during the training, we are able to provide models that do not need these information during inference stage, thus making it a lot faster, as estimating dense optical flow is operationally expensive (time and computational wise). 

The content loss and the style loss utilize feature maps of the VGG network at the relu2\_2 layer and [relu1\_2, relu2\_2, relu3\_3, relu4\_3] layers respectively. We used VGG16 pre-trained in ImageNet for calculating all of the losses.





\section{Experiments and Discussion}
\label{sec:experiments}


For the training of the network, we used Coco \cite{cocoDataset} and MPI Sintel \cite{Mpi_sintel}  datasets. All image frames were resized to $256 \times 256$ pixels for the first stage of the training. In the second phase we used MPI Sintel, which required $640\times360$ resolution. We implement our style transfer pipeline on PyTorch 1.0.1  with Cuda 10.1 and cuDNN 7.1.4. All of the experiments were performed on a single GTX 1080 GPU. In the next subsections, we will discuss the differences between deploying DNN on the Android and iPhone system .

\begin{figure}[ht!]
\caption{Content image for experiments presented in the Fig. \ref{fig:differentStyle}}
\label{different_Style_original}
\centering
\includegraphics[width=0.49\textwidth]{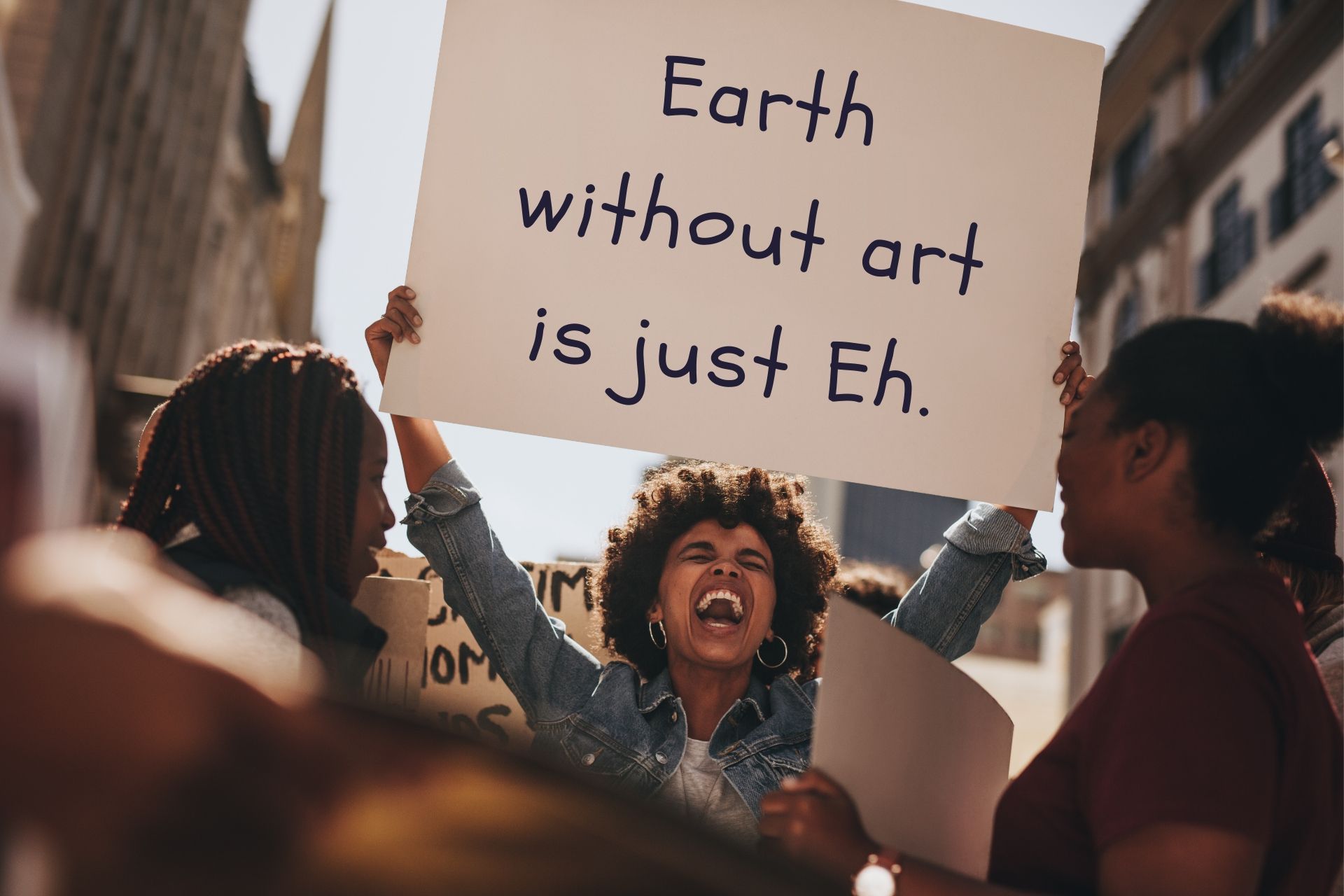}
\end{figure}

\begin{figure*}[ht]
\begin{tabular}{cccc}

\hline
a) Scream  & b) Mona Lisa & c) Weeping Woman & d) Starry Night \\
\hline
\IncG[width=.21\textwidth,height=.23\textwidth]{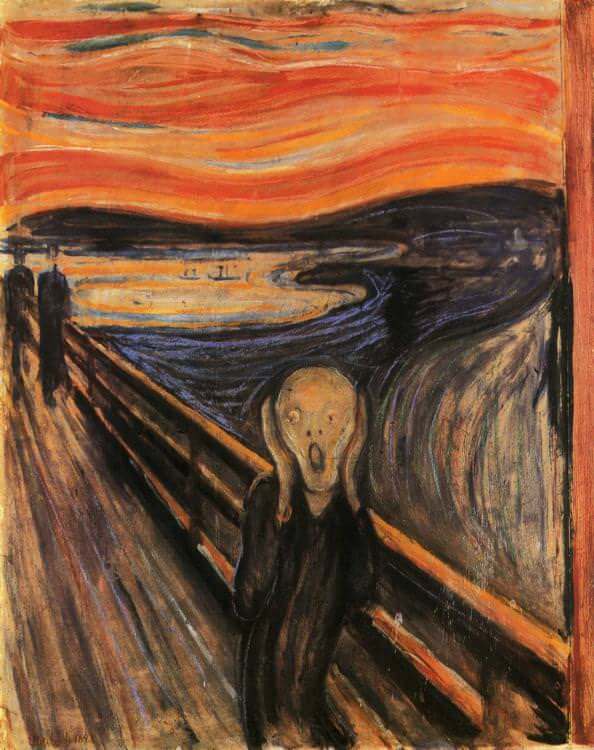} &
\IncG[width=.18\textwidth,height=.23\textwidth]{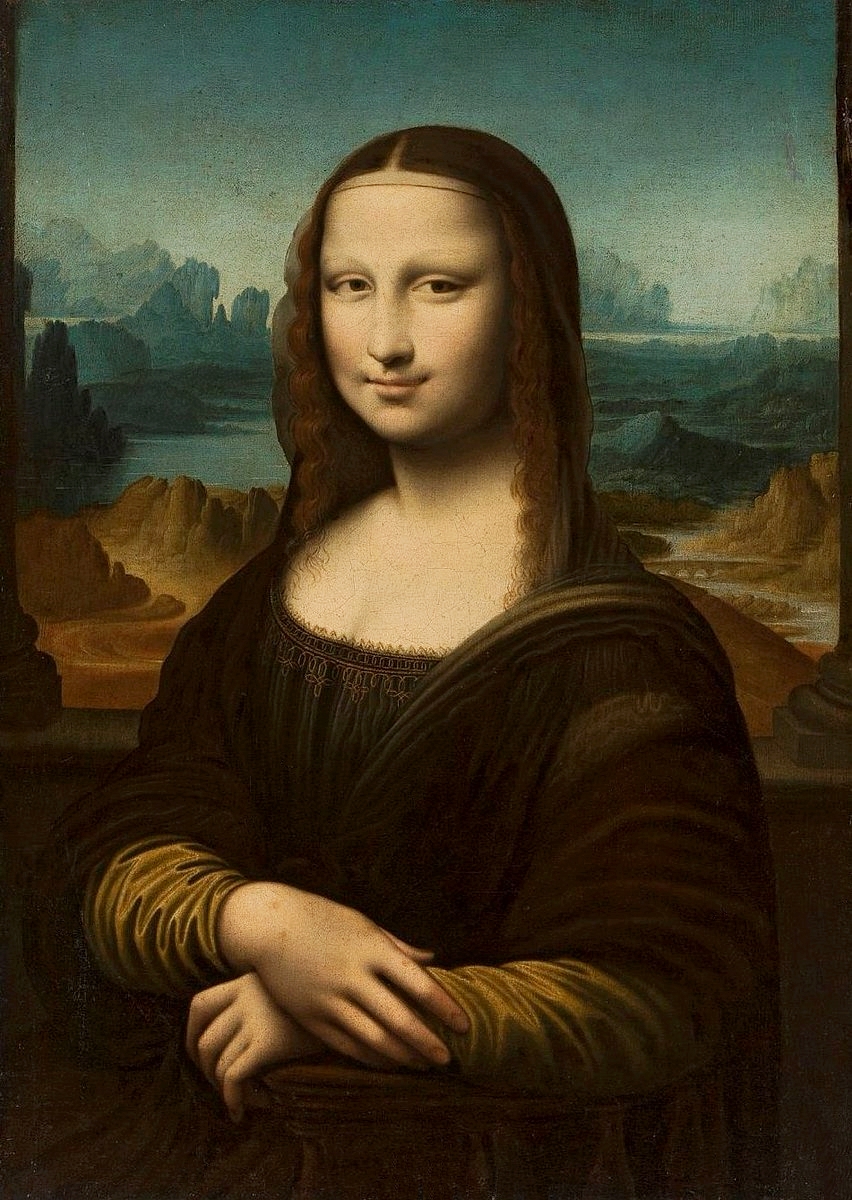} &
\IncG[width=.21\textwidth,height=.23\textwidth]{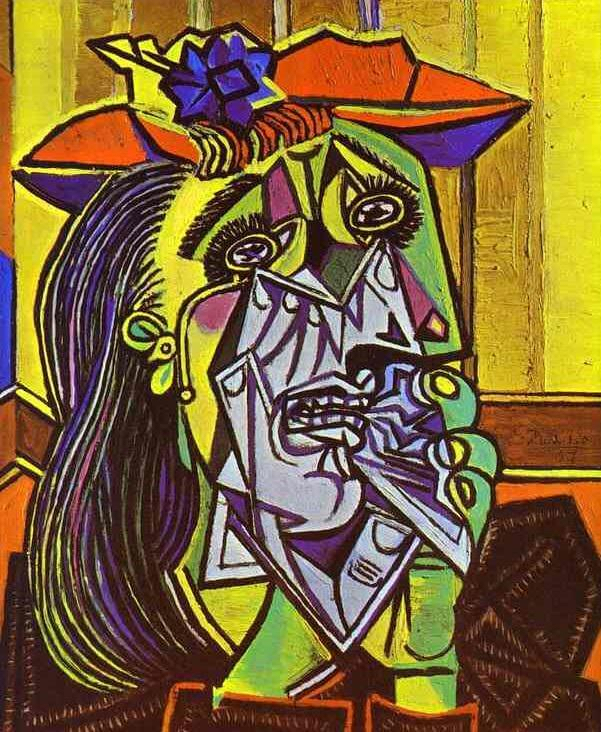} &
\IncG[width=.23\textwidth,height=.21\textwidth]{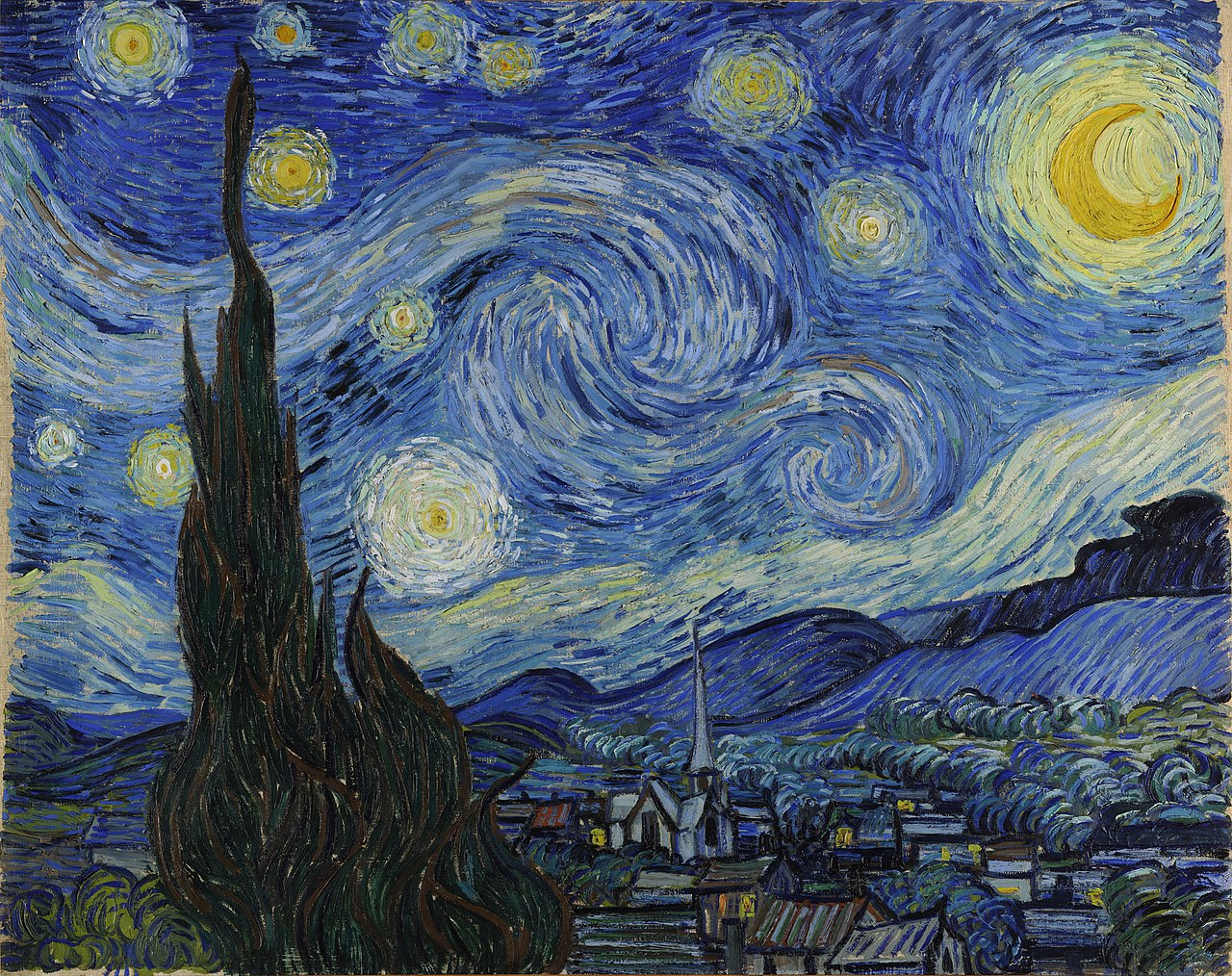}
\\

\IncG[width=.23\textwidth,height=.18\textwidth]{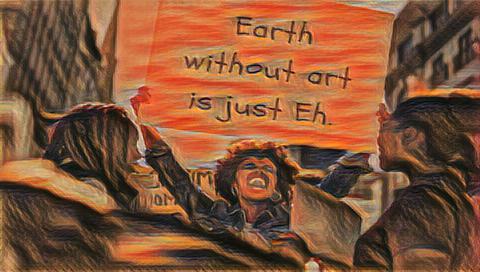} &
\IncG[width=.23\textwidth,height=.18\textwidth]{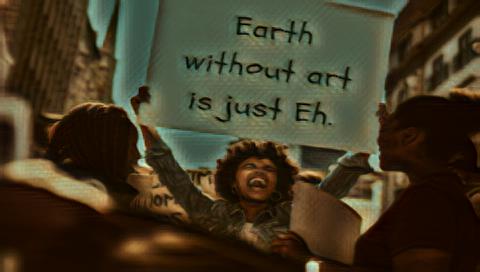} &
\IncG[width=.23\textwidth,height=.18\textwidth]{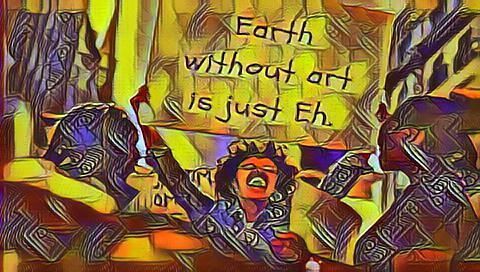} &
\IncG[width=.23\textwidth,height=.18\textwidth]{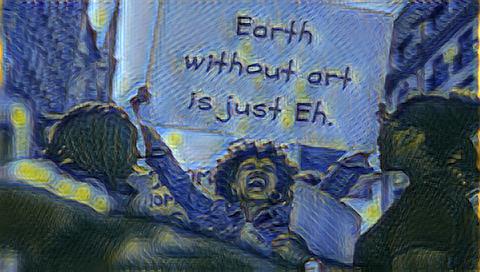}
\\

\IncG[width=.23\textwidth,height=.18\textwidth]{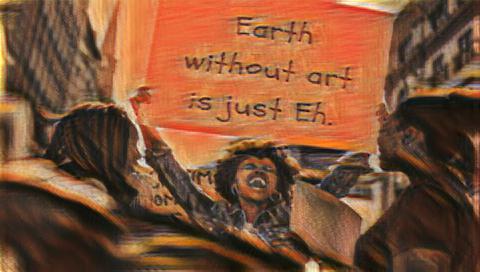} &
\IncG[width=.23\textwidth,height=.18\textwidth]{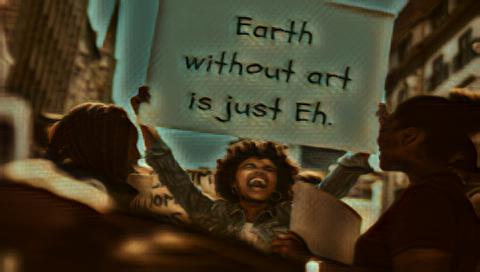} &
\IncG[width=.23\textwidth,height=.18\textwidth]{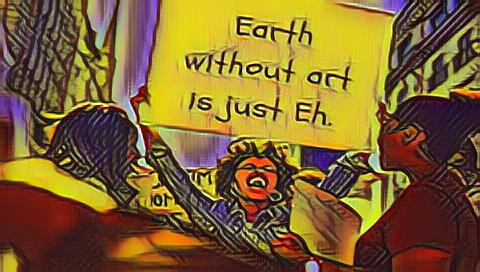} &
\IncG[width=.23\textwidth,height=.18\textwidth]{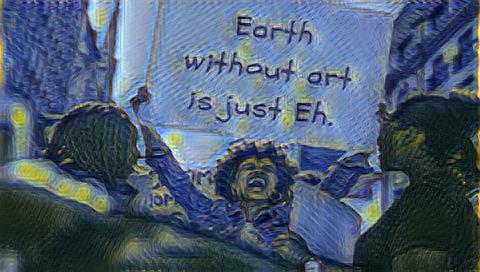}
\\
\hline

\end{tabular}
\caption{Examples of different stylization achieved with the presented training technique. First row is style image, second is result after first phase of training, the last row presents final model's results. }
\label{fig:differentStyle}
\end{figure*}

The results of the two-stage training procedure is depicted in the Fig. \ref{fig:differentStyle}, while the content image is presented in the Fig. \ref{different_Style_original}. The first row presents the style images, while the second row we presents the result of the model after the first phase of training. In the third row, we present the same images after incorporating fine-tuning for stabilization of the video. As we can see, especially in the example of Weeping Woman and Scream, introducing temporal coherence into the model weakened the style that was learned by the neural network. The weakening of the style is mainly visible as smoothing out the resulting image so the colors are more unified across a wide area of the image. As presented, achieving the desired stabilization introduces some trade-off in the case of some particular styles. As an advantage over other methods, our two-stage training approach can provide both models (with and without stabilization) thus letting the author decide what effect is desired.

\subsection{iOS vs Android implementation}
\label{Andorid_and_ios}

Deployment of DNN on mobile devices introduces many difficulties. These difficulties include, among other things: hardware capabilities, proper drivers for hardware, and deep learning frameworks implementation. What is more, there are some noticeable differences between Android and iOS devices. First of all, Apple introduced a dedicated NPU with the A11 Bionic system on chip (SoC) back in 2017. Although not available to third-party applications, the next-generation A12 SoC changed that in 2018. This allows for very fast and power-efficient calculations with the dedicated NPU. What is more thanks to the Apple ecosystem, they were able to provide the CoreML library that hides the technical details from the developer. During multiple numerical experiments, we have noticed that our neural network implementation worked fast on Apple devices without a dedicated NPU. This was possible due to the fact that CoreML library can also exploit the GPU. The switch between NPU and GPU on different devices is done automatically, depending on their hardware. Another advantage of the CoreML, is that it supports a wide variety of formats as well as provides a proper conversion mechanism for different models e.g ONNX $\rightarrow$ CoreML, Keras $\rightarrow$ CoreML.

The Android devices are not so unified which brings a number of problems during development. The first to follow, there is no single library to cover multiple use-cases. In our opinion, most promising solution for DNN projects is Tensorflow Lite, however this may change in the future.  Being relatively  mature library, TF Lite,  it is not a silver bullet. Depending on the backend used (CPU, GPU or dedicated hardware) different operations (layers) might not be supported. It may depend on things such as: kernel size, stride value, or padding options. These dependencies involves an automatic mechanism of switching computation from the desired hardware to the CPU when operation is not supported (CPU is often the slowest possible option). These situations must be carefully examined and addressed. Another problem when using Tensorflow Lite is the conversions of models trained using different frameworks. Some of the conversion tools that exists might introduce additional layers. A great example can be shown in converting Pytorch to Tensorflow through the common format of ONNX. As these frameworks depends on a different layout of data (Tensorflow uses NHWC order of data while Pytorch uses NCHW), any kind of conversion adds transform layers which significantly impact the performance. Numerical experiments with conversion (not published) showed, that our application noted a 30-40\% drop in the frames per second.

Another library worth mentioning is Pytorch, which with the version 1.3, allowed the execution of the models on mobile devices. While this is s promising development direction, it is still lacking the GPU and Andorid Neural Network API (NNAPI) support,  which is currently the major drawback. The variety of Android devices results in another problem. There are multiple device manufactures and chip manufacturers, which causes slow adoption of the NNAPI. What is more the usage of NNAPI depends on the drivers provided by the device manufacturer. With no drivers provided, it defaults to the use of CPU. On the other hand, there might be major differences between smartphones models in terms of supporting the NNAPI as some of them might support only quantized (INT8) or float (FP32) only models. All of that make it hard to predict how a deployed neural network model will behave across multiple devices. As mentioned in Ignatov et al\cite{aiMobileBenchmark}, the newest generation of NPU seems to provide a wide set of commonly supported operations. These NPUs supports both quantized and float models at the same time. All of that shows a great progress made in recent years on the Android device which should make it easier to deploy deep learning model on both of the platforms, as until now, Android was inferior in comparison with iOS.


\subsection{Investigation on size of network}


\begin{table*}[]
\centering
\scriptsize
\caption{Configurations tested}
\label{tab:my-table}
\begin{tabular}{rrrrrrrrrrr}
\hline
\multicolumn{1}{c}{\multirow{2}{*}{Id}} & \multirow{2}{*}{$\alpha$} & \multirow{2}{*}{$\beta$} & \multirow{2}{*}{parameters} & \multirow{2}{*}{\% of ReCoNet} & \multirow{2}{*}{Size (MB)} & \multirow{2}{*}{\% of ReCoNet} & \multicolumn{2}{c}{iPhone 11 Pro} & \multicolumn{2}{c}{iPhone 6s} \\
\multicolumn{1}{c}{}                    &                        &                       &                             &                                &                            &                                & FPS ($480\times320$)     & ($320\times 240$)  & FPS ($480\times320$)   & FPS ($320\times 240$)   \\ \hline
1                                       & 1.000                  & 1                     & 470k                        & 15.16                          & 1.79                       & 15.14                          & 12.26           & 19.57           & 4.96          & 9.48          \\
2                                       & 0.750                  & 1                     & 267k                        & 8.64                           & 1.02                       & 8.63                           & 14.91           & 21.41           & 6.13          & 13.65         \\
3                                       & 0.500                  & 1                     & 122k                        & 3.94                           & 0.47                       & 3.98                           & 21.91           & 25.17           & 13.27         & 19.45         \\
4                                       & 0.250                  & 1                     & 33k                         & 1.06                           & 0.12                       & 1.02                           & 27.72           & 40.91           & 19.33         & 24.16         \\
5                                       & 0.125                  & 1                     & 9k                          & 0.30                           & 0.04                       & 0.34                           & 34.08           & 46.41           & 22.77         & 33.98         \\ \hline
6                                       & 1.000                  & 0.75                  & 307k                        & 9.93                           & 1.17                       & 9.90                           & 15.1            & 18.53           & 8.49          & 12.93         \\
7                                       & 0.750                  & 0.75                  & 173k                        & 5.61                           & 0.66                       & 5.58                           & 16.66           & 21.52           & 8.56          & 17.65         \\
8                                       & 0.500                  & 0.75                  & 78k                         & 2.51                           & 0.30                       & 2.54                           & 21.28           & 34.77           & 18.29         & 21.73         \\
9                                       & 0.250                  & 0.75                  & 20k                         & 0.64                           & 0.08                       & 0.68                           & 38.63           & 51.37           & 22.82         & 35.03         \\
10                                      & 0.125                  & 0.75                  & 5k                          & 0.17                           & 0.02                       & 0.17                           & 49.92           & 60.41           & 33.75         & 48.38         \\ \hline
11                                      & 1.000                  & 0.5                   & 233k                        & 7.54                           & 0.89                       & 7.53                           & 15.66           & 18.93           & 9.25          & 14.05         \\
12                                      & 0.750                  & 0.5                   & 131k                        & 4.26                           & 0.50                       & 4.23                           & 18.21           & 23.43           & 9.40          & 18.65         \\
13                                      & 0.500                  & 0.5                   & 59k                         & 1.91                           & 0.23                       & 1.95                           & 22.64           & 37.68           & 19.07         & 23.21         \\
14                                      & 0.250                  & 0.5                   & 15k                         & 0.49                           & 0.06                       & 0.51                           & 40.01           & 53.90            & 24.19         & 38.16         \\
15                                      & 0.125                  & 0.5                   & 4k                          & 0.13                           & 0.02                       & 0.17                           & 51.92           & 62.43           & 35.93         & 50.56         \\ \hline

\end{tabular}
\end{table*}

\begin{figure}[ht!]
\caption{A graph showing FPS versus number of parameters (in thousands) in DNN model tested with iPhone 11 Pro and iPhone 6s. The x-axis is presented in logarithmic scale}
\label{graphFPS}
\centering
\includegraphics[width=0.49\textwidth]{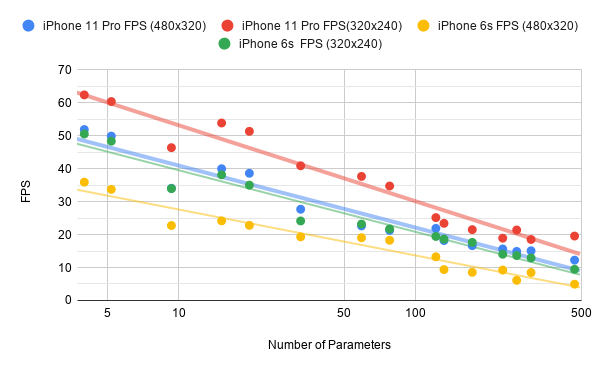}
\end{figure}

\begin{figure}[ht!]
\begin{tabular}{cc}

\IncG[width=.25\textwidth,height=.2\textwidth]{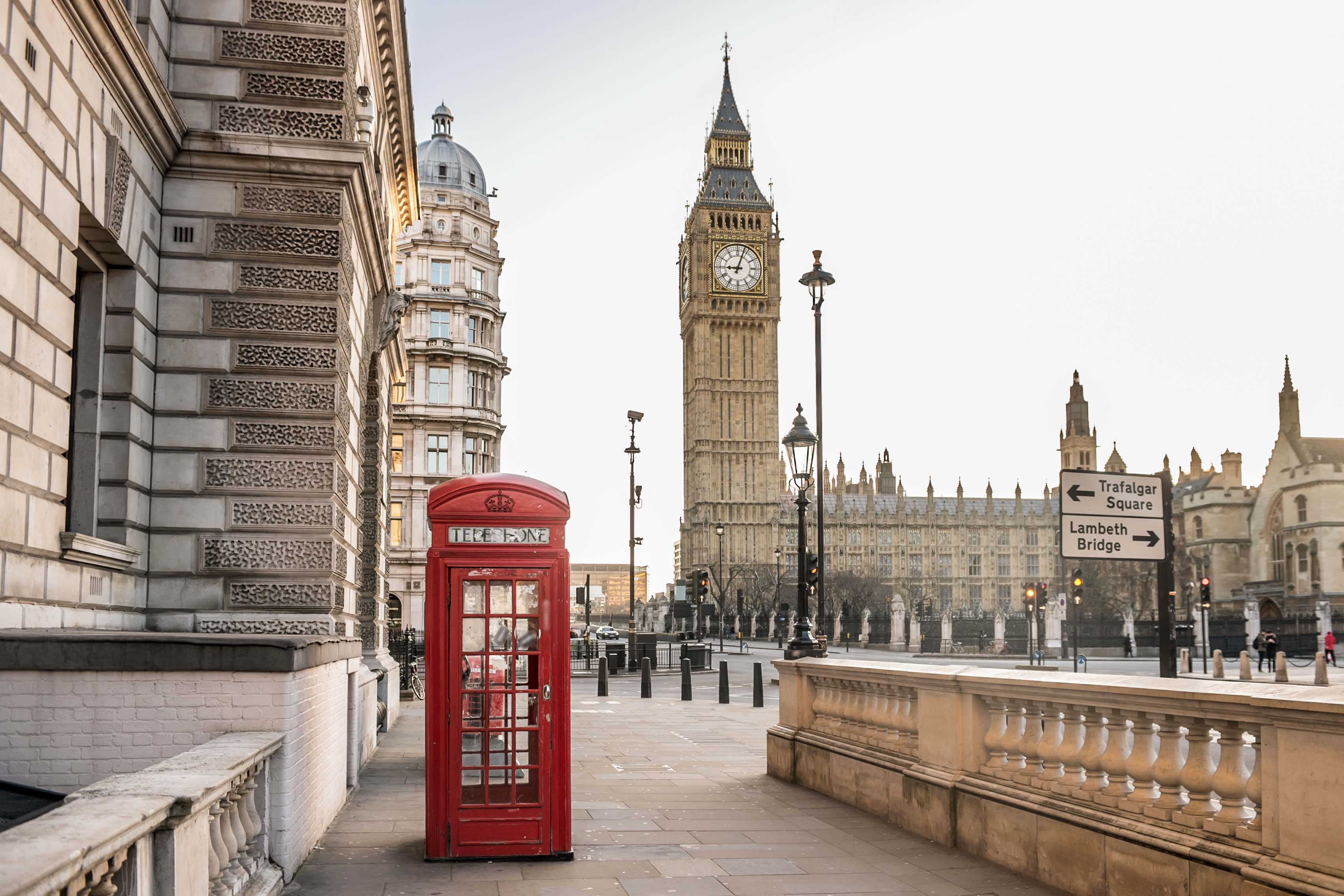} &
\IncG[width=.2\textwidth,height=.2\textwidth]{figures/style/scream.jpg} 
\\
a) Content image  & b) Style image 
\end{tabular}
\caption{ Content and style images for experiments in Fig. \ref{fig:comparison}}
\label{fig:content_and_style}
\end{figure}

\begin{figure*}[ht!]
\begin{tabular}{cccc}

\hline
$\alpha$ &   \multicolumn{3}{c}{$\beta$} 
\\
\hline
& \multicolumn{1}{c}{1} & \multicolumn{1}{c}{0.75}   & \multicolumn{1}{c}{0.5}  \\

\rotatebox[origin=c]{90}{1} &
\IncG[width=.3\textwidth,height=.22\textwidth]{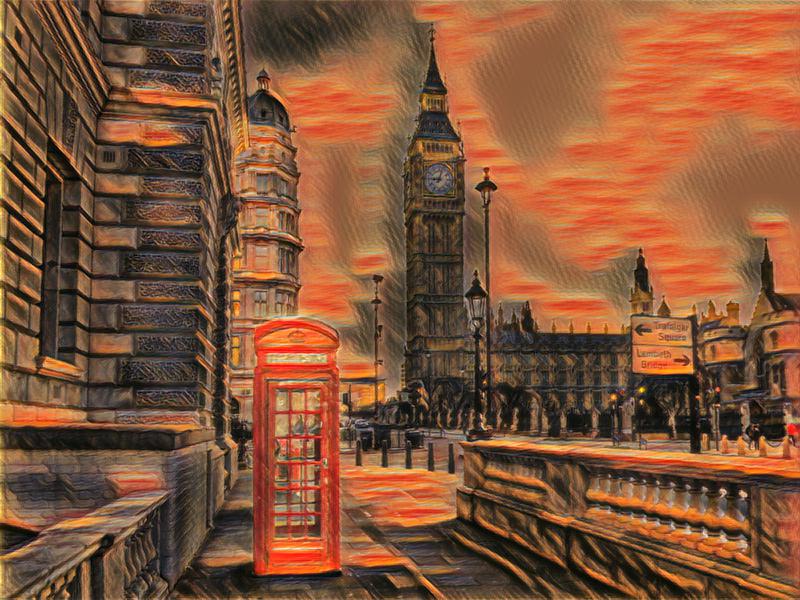} &
\IncG[width=.3\textwidth,height=.22\textwidth]{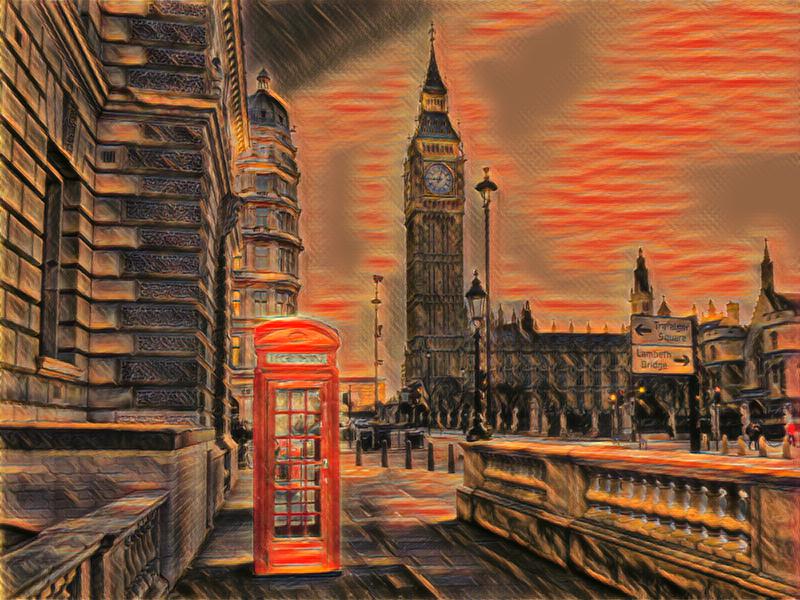} &
\IncG[width=.3\textwidth,height=.22\textwidth]{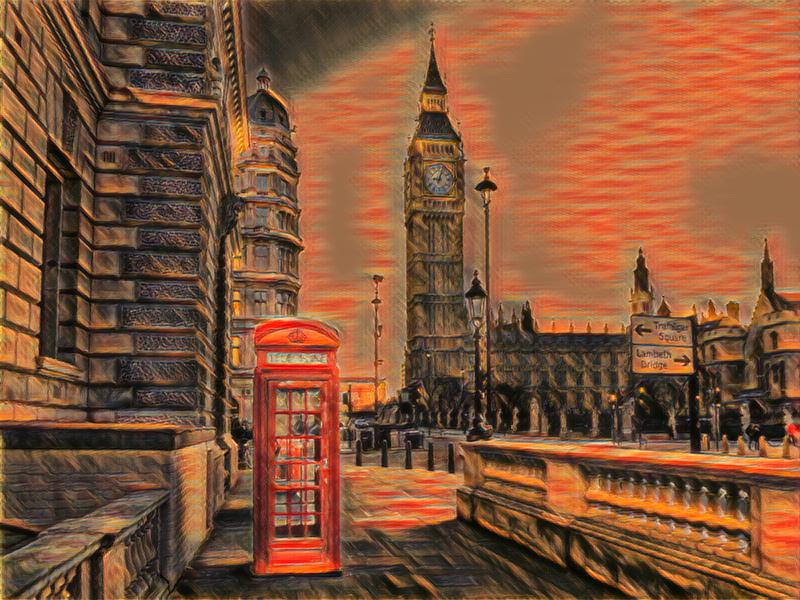} 
\\

\rotatebox[origin=c]{90}{0.75} &
\IncG[width=.3\textwidth,height=.22\textwidth]{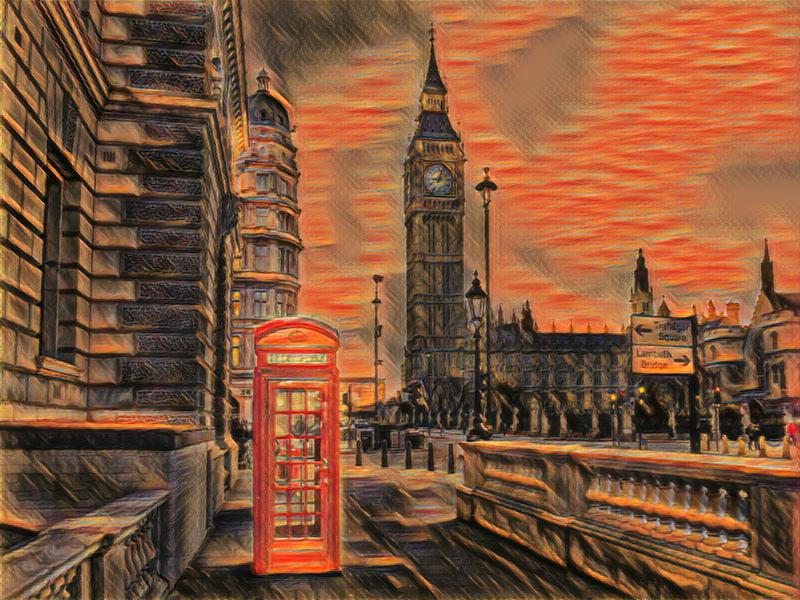} &
\IncG[width=.3\textwidth,height=.22\textwidth]{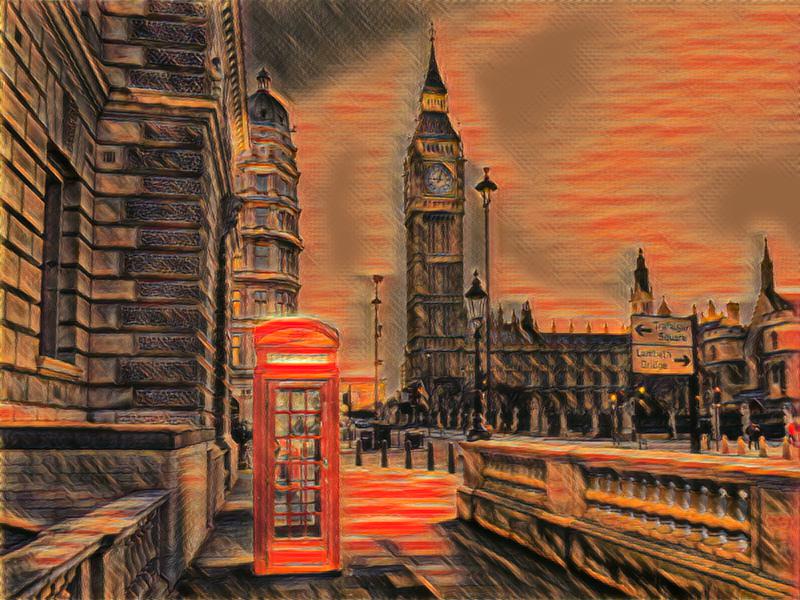} &
\IncG[width=.3\textwidth,height=.22\textwidth]{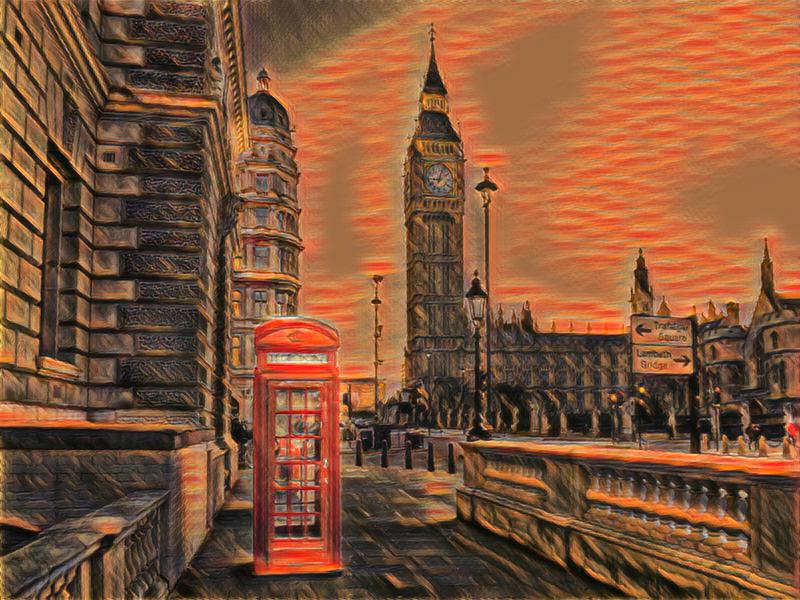}
\\

\rotatebox[origin=c]{90}{0.5} &
\IncG[width=.3\textwidth,height=.22\textwidth]{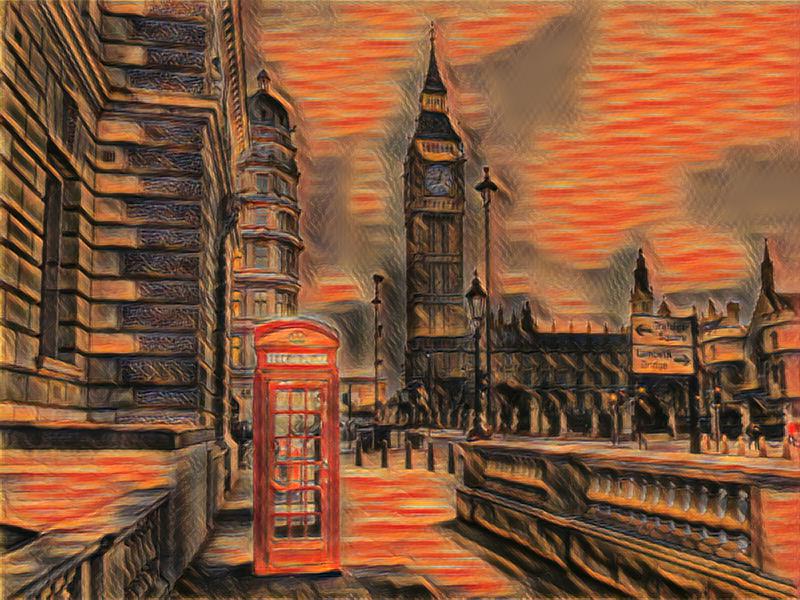} &
\IncG[width=.3\textwidth,height=.22\textwidth]{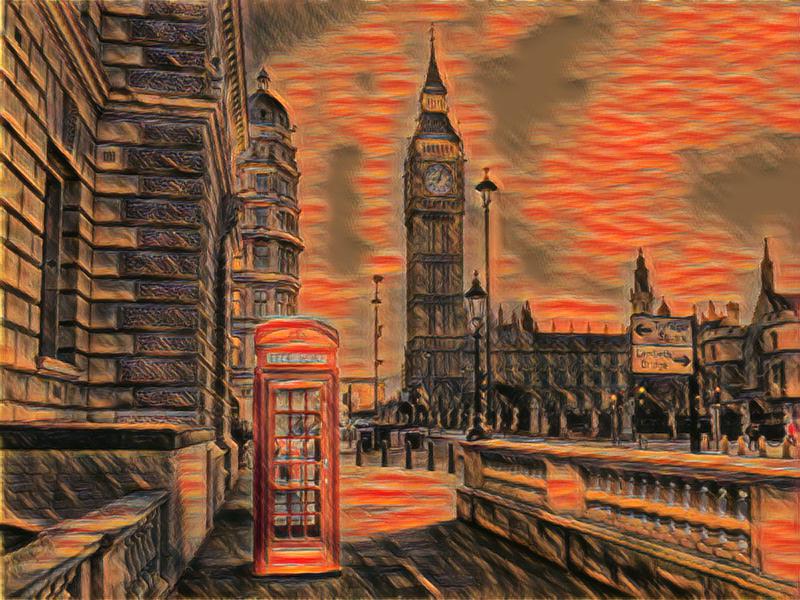} &
\IncG[width=.3\textwidth,height=.22\textwidth]{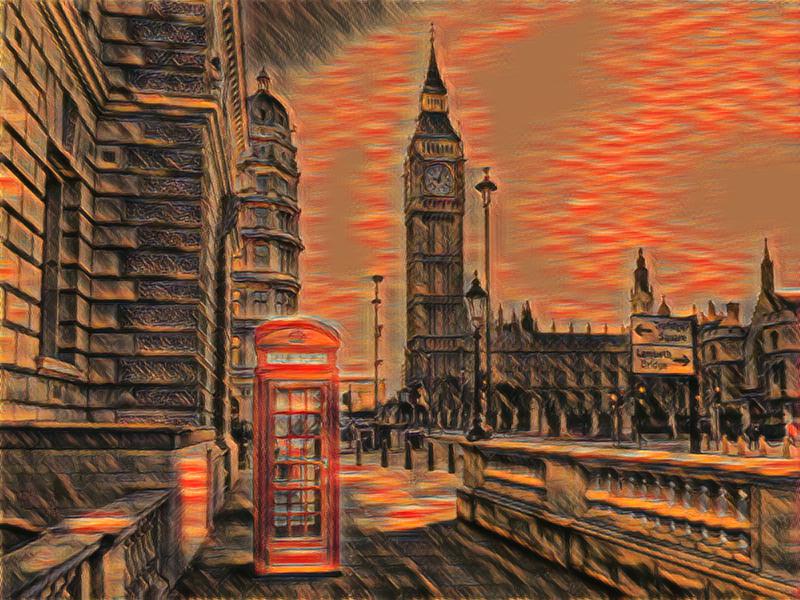} 
\\

\rotatebox[origin=c]{90}{0.25} &
\IncG[width=.3\textwidth,height=.22\textwidth]{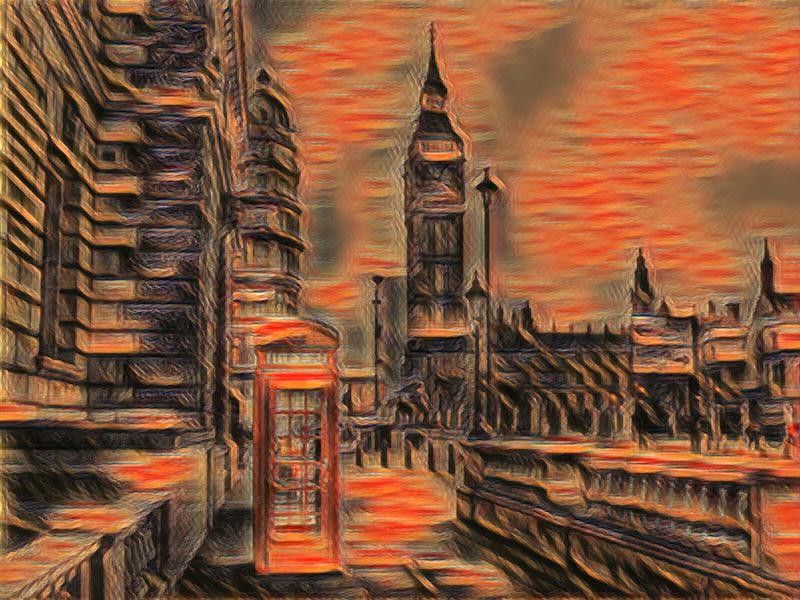} &
\IncG[width=.3\textwidth,height=.22\textwidth]{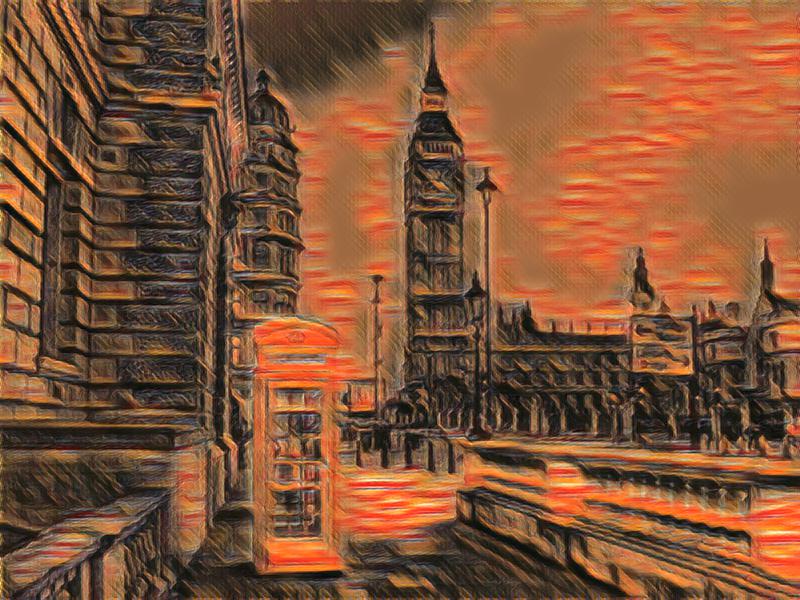} &
\IncG[width=.3\textwidth,height=.22\textwidth]{figures/224_ben_800_600.jpg} 
\\

\rotatebox[origin=c]{90}{0.125} &
\IncG[width=.3\textwidth,height=.22\textwidth]{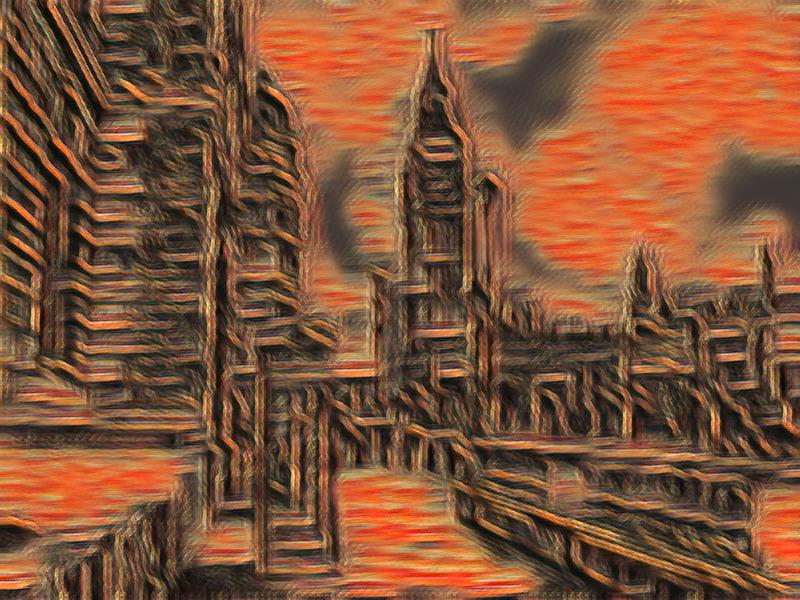} &
\IncG[width=.3\textwidth,height=.22\textwidth]{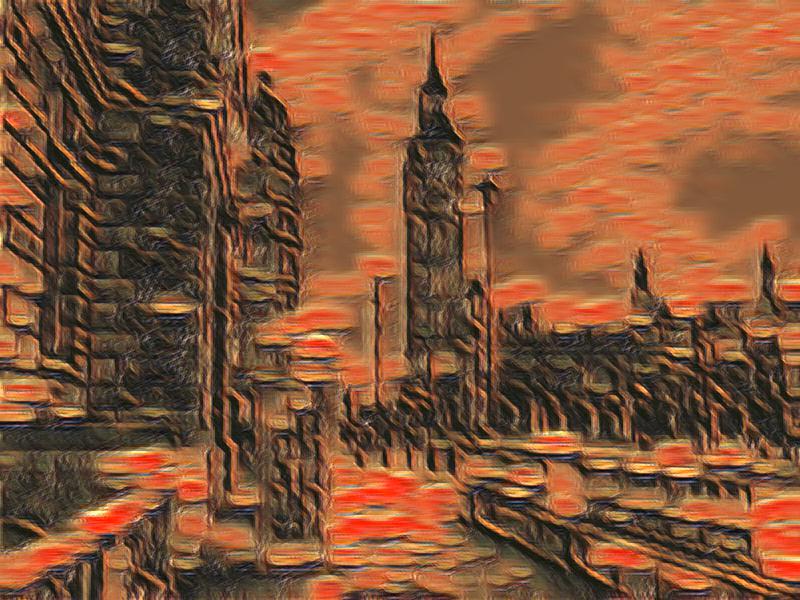} &
\IncG[width=.3\textwidth,height=.22\textwidth]{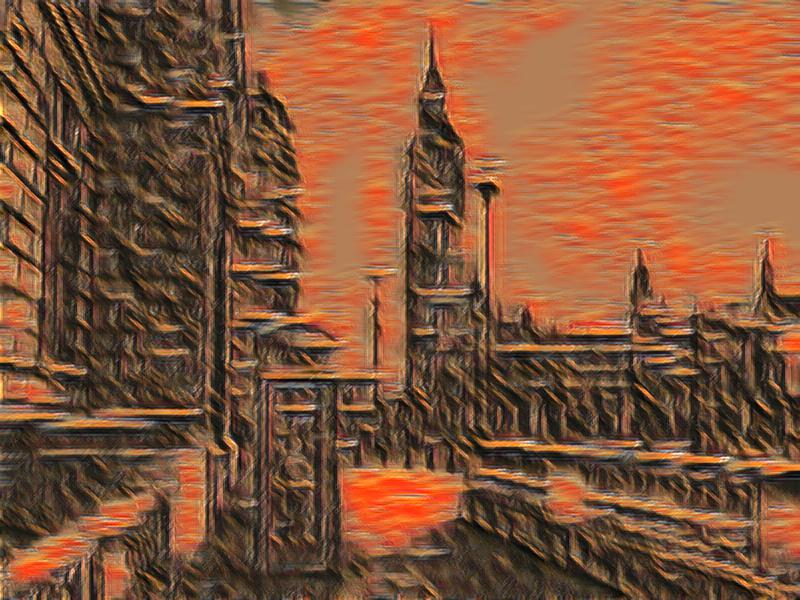} 
\\
\hline
\end{tabular}
\caption{ Qualitative comparison of stylization performed by networks with different size}
\label{fig:comparison}
\end{figure*}

In Tab. \ref{tableWithArchitecture}, we introduced $\alpha$ and $\ResidualLayerNumber$  parameters for the number of filters and number of residual layers.  We ran training sessions with these parameters, to check how they impact achieved style transfer. In the Tab. \ref{tab:my-table}, we present the results of our experiments. The qualitative (visual) comparison for the network trained with "Scream" by Munch as style image is presented in the Fig. \ref{fig:comparison} while the content and style image for these models are presented in Fig. \ref{fig:content_and_style}. All of the networks were trained with the exactly the same settings for hyperparameters.

We also did preliminary experiments with the quantization of the networks to INT8 format but we did not notice a significant improvement in terms of processing speed (frames per second). As our networks run on mobile GPU, it might happen that there is no special hardware to exploit the quantization. As a result, the calculation with INT8 or FLOAT32 take a similar amount of time.  While this is  an uncommon situation, we need to investigate it further because this might depend on certain device models used for those experiments. While the pruning of an already trained network can be another solution to finding the optimal size of the networ,k we did not have a chance to investigate it deeply enough. The problem that we see for these techniques is that the value of loss function for style transfer is not showing the final results that the network will provide. This makes it hard to assess the neurons that should be deleted during the pruning with the restriction of providing a similar result to the original network. The best way to measure tha,t is the qualitative approach of comparing two images which might be subjective and opinion-based. Moreover, pruning tends to produce sparse networks (with sparse matrices)  which may not be faster than the full network. Also as presented in \cite{pruning_Qiangui} pruning mostly happens in fully connected layers which we do not use in our architecture.

As we can see in Fig. \ref{fig:comparison}, the smallest networks (id: 5, 10 and 15) provided very poor results, with multiple deformations which made it hard to recognize the original content image. At the same time, networks denoted with id 4, 9 and 14 provide much better results and are able to capture both: style and content. In our opinion, these results indicate that there is a tradeoff between the keeping good-looking stylization and remaining proper content. What is interesting we see that the model 14 is able to provide the most fine details (look at the clock tower of Big Ben) from those three while being the smallest, however, this type of evaluation is purely subjective. We can also hypothesize  that the model capacity of all of these networks id 4, 9, and 14 should be sufficient for learning the style. However, achieving subjectively good results may depend on the initialization of network parameters. In some  cases, it may be difficult to recognize the differences between the rest of the models. As there are some clear differences visible between them, in our opinion the style and content is represented with the same level of detail and choosing one over another would be a matter of individual preferences about the certain style. 

No surprise, the reduction of the model's size on disk is proportional to the reduction of the number of parameters . It is well-known that it is the parameters of networks that take the most space in the saved model (see Tab. \ref{tableWithArchitecture}). We are also showing the reduction in both of those terms comparing it to ReCoNet \cite{Reconet}. The size of the network is an important matter when deploying it to mobile devices as the size of the application may depend heavily on that. When each network provides a single artist style, having multiple styles in a single application can quickly increase its size. The original ReCoNet network size is $11.82$ MB while our experiments ranges form $1.79$ to $0.02$ MB. Thus we are able to provide multiple models while keeping a smaller memory footprint. The performance of our model in terms of frames per second (FPS) was measured on the iPhone 11 Pro and iPhone 6s with two different resolutions of the input for model. The measurement was taken by applying the model to the video with a resolution of $1920\times1080$, length of $21$, seconds coded with H.264 and 25 FPS. The results present in Tab. \ref{tableWithArchitecture} are the average speed of each model after processing this video (resizing the image to network input size is included into this measurement). As we can see, decreasing the number of parameters for the network gives non-linear growth in the speed of processing. This trend is presented in the Fig. \ref{graphFPS}. We can see the logarithmic relationship between the number of FPS and number of parameters. The difference between the models  1 and  2 is around $200k$ parameters while this shrinking provides only 2 more FPS (for both tested devices) on averaged in favor of the smaller model. The further shrinking of models (id 3, 8, and 13) provides near real-time performance with 22 FPS on the $480\times320$ resolution  and real-time performance (over 25 FPS) with $320\times240$ resolution for the iPhone 11 Pro. In the case of the iPhone 6s device, smaller models, id 4, 9, and 14, are needed to achieve similar performance. While we are not aware of the hardware design of the Apple chips, what can be seen is that removing residual layers has rather a small impact on the overall performance of the model while $\alpha$ is much more important. This is true for both of the tested devices. This performance impact can be especially be seen when comparing models with $\alpha=1.0$ and different $\ResidualLayerNumber$. Despite a great overall difference in the number of parameters between them, the final performance stays is similar for $320\times240$ resolution and the iPhone 11 Pro device. In this case, we can see the fluctuation of measurement where smaller models show decreased performance as we were simulating a real use case with some applications running in background. To conclude, we have proven that the size of the network, in terms of number of parameters and the model size on the disk, can be greatly reduced and we are able to provide real-time performance on mobile devices without losing the quality of the expected result.

\section{Conclusions}

In this paper, we presented a feed-forward neural network based on ReCoNet \cite{Reconet} for real-time video style transfer on mobile devices. We showed technical challenges in deploying CNN to mobile devices and the differences between the Android and iPhone ecosystems with recommendations for any future work. This includes that we need to be very careful about using certain layers and kernel sizes in order to achieve expected performance. We also proposed a novel way of achieving temporal consistency by re-training already available models. By this, we made it easy to use models that are trained with other methods. At last, we also investigated the network size regarding the number of filters and number of residual layers and showed that shrinking them to less than $4\%$ of the number of parameters present in \cite{Reconet} we are still able to achieve good looking results as shown by our experiments. The results of that work can be checked with the Kunster -- AR Art Video Maker mobile application on iOS devices.

\bibliographystyle{IEEEtran}
\bibliography{ms}

\end{document}